\pdfoutput=1

\documentclass[11pt]{article}

\usepackage[final]{acl}

\usepackage{times}
\usepackage{latexsym}

\usepackage[T1]{fontenc}

\usepackage[utf8]{inputenc}

\usepackage{microtype}

\usepackage{inconsolata}

\usepackage{hyperref}
\usepackage{url}
\usepackage{subcaption}
\usepackage{booktabs}
\usepackage{multirow}
\usepackage{xcolor}
\usepackage{marginnote}
\usepackage{amsfonts}
\usepackage{graphicx}
\usepackage{wrapfig}
\usepackage{amsmath}

\usepackage{algorithm}
\usepackage{algorithmic}

\title{Advancing Adversarial Suffix Transfer Learning on Aligned Large Language Models}

\author{
 \textbf{Hongfu Liu\thanks{\ \ Equal contribution.}},
 \textbf{Yuxi Xie\footnotemark[1]},
 \textbf{Ye Wang},
 \textbf{Michael Shieh}
\\
 National University of Singapore \\
 \texttt{\{hongfu,wangye,michaelshieh\}@comp.nus.edu.sg, xieyuxi@u.nus.edu}
}

\begin{document}
\maketitle
\begin{abstract}

Language Language Models (LLMs) face safety concerns due to potential misuse by malicious users. Recent red-teaming efforts have identified adversarial suffixes capable of jailbreaking LLMs using the gradient-based search algorithm Greedy Coordinate Gradient (GCG). However, GCG struggles with computational inefficiency, limiting further investigations regarding suffix transferability and scalability across models and data. In this work, we bridge the connection between search efficiency and suffix transferability. We propose a two-stage transfer learning framework, DeGCG, which decouples the search process into behavior-agnostic pre-searching and behavior-relevant post-searching. Specifically, we employ direct first target token optimization in pre-searching to facilitate the search process. We apply our approach to cross-model, cross-data, and self-transfer scenarios. Furthermore, we introduce an interleaved variant of our approach, i-DeGCG, which iteratively leverages self-transferability to accelerate the search process. Experiments on HarmBench demonstrate the efficiency of our approach across various models and domains. Notably, our i-DeGCG outperforms the baseline on Llama2-chat-7b with ASRs of $43.9$ ($+22.2$) and $39.0$ ($+19.5$) on valid and test sets, respectively. Further analysis on cross-model transfer indicates the pivotal role of first target token optimization in leveraging suffix transferability for efficient searching\footnote{Code is publicly available at \href{https://github.com/Waffle-Liu/DeGCG}{https://github.com/Waffle-Liu/DeGCG}}.


\end{abstract}

\section{Introduction}





















Large Language Models (LLMs) have become integral to everyday decision-making processes~\citep{gpt4,google,llama2}. However, alongside the convenience they offer, there is increasing concern about their potential to produce harmful and ethically problematic responses to user queries, which raises significant safety issues. In response to these concerns, recent efforts have focused on aligning LLMs with human preferences to enhance the responsibility and harmlessness of their responses~\citep{align3,align1,align2}. Despite these alignment efforts, LLMs still remain vulnerable to potential attacks~\citep{jailbroken}. Recent studies have revealed various types of jailbreak attacks~\citep{jailbroken,albert2023jailbreakchat,kang2023exploiting,sesame,autodan}, which involve using jailbreak prompts alongside malicious queries to compel aligned LLMs to generate harmful and unethical responses, thereby circumventing the safety alignment constraints. 



 \begin{figure}[t]
    \centering
    \includegraphics[scale=0.37]{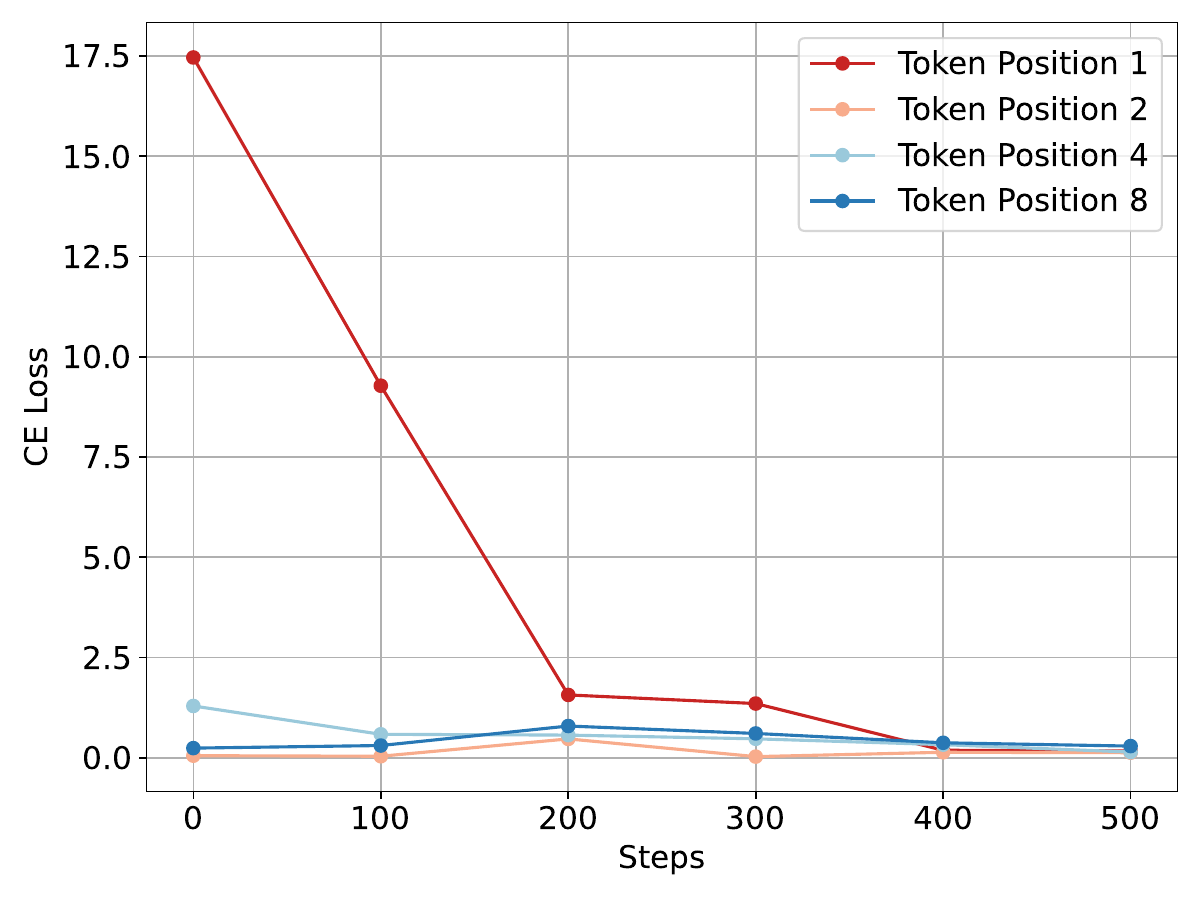}
    \caption{GCG Training Dynamics of Cross Entropy Loss for tokens located at different positions in the target sequence. We plot the changes in cross-entropy loss of target tokens at positions [1, 2, 4, 8] every 100 steps. This discrepancy in loss dynamics highlights the importance of first token optimization in GCG.}
\label{fig:token-level-ce}
\vspace{-5mm}
\end{figure}

One notable attack, Greedy Coordinate Gradient (GCG)~\citep{gcg}, utilizes gradient information to search for adversarial prompts, also known as adversarial suffixes, which can be appended to malicious queries to elicit harmful responses. These adversarial suffixes consist of random tokens and are generally not comprehensible to humans. However, deriving these suffixes through gradient-based searching is computationally inefficient. The exponentially increasing search space of random suffixes with length expansion presents significant challenges to search efficiency. Besides, the random initialization for each search is inefficient, incurring additional but unnecessary searching costs. Recent work~\citep{gcg} suggests that the adversarial suffixes may possess universal transferability across models, indicating that the previously searched suffix could serve as an effective initialization. Furthermore, \citet{meade2024universal} finds that models aligned through preference optimization exhibit robustness against suffix transfer. Despite these insights, prior works primarily focused on direct transfer, which shows limited transferability across different models or data domains. The potential for using adversarial suffixes as initialization for transfer learning remains largely unexplored.


In this work, motivated by the challenges in optimizing the gradient-based search process with effective initial adversarial suffixes, we explore how to leverage the transferability of these suffixes during optimization. Our empirical investigation has identified the importance of optimizing the first target token loss, as illustrated in Fig.~\ref{fig:token-level-ce}. We attribute the inefficiency in searching to the cross-entropy optimization goal applied to the entire target sentence. To address this, we propose a two-stage transfer learning framework, DeGCG, which decouples the original search process into two stages: behavior-agnostic pre-searching and behavior-relevant post-searching: 
\begin{itemize}
    \item In the pre-searching stage, we perform a simplified task, First-Token Searching (FTS), searching for adversarial suffixes with a behavior-agnostic target such as ``Sure'', enabling LLMs to elicit the first target token without refusal.
    \item In the post-searching stage, we start with the suffix obtained from the pre-searching stage and conduct Content-Aware Searching (CAS) with a behavior-relevant target. This stage transfers the behavior-agnostic initialization to behavior-relevant suffixes.
\end{itemize}

We found that suffixes obtained through first-token searching can be effectively transferred across different models and datasets with further searching. Additionally, we leverage the self-transferability of adversarial suffixes and propose an interleaved training algorithm, i-DeGCG, which performs FTS and CAS in an interleaved manner. We evaluate our proposed method on the HarmBench across various LLMs. Our experimental results demonstrate the effectiveness and efficiency of the DeGCG framework and i-DeGCG variant, highlighting the success of suffix transfer through two-stage learning and underscoring the importance of initialization for search efficiency.
  



\section{Related Work}

\subsection{Safety-Aligned LLMs}
LLMs have demonstrated impressive capabilities but raised safety concerns about the potential for malicious usage. To mitigate these concerns, efforts have been made to supervised fine-tuning of LLMs with instructions aimed at ensuring helpfulness and safety~\citep{instruction,inst-tuning,llama2}, and align LLMs with human preference, known as Reinforcement Learning from Human Feedback (RLHF)~\citep{rlhf,align1,align2,align3}. RLHF involves training LLMs based on the rewards derived from models that have been trained on human preference data. Recent studies show that models aligned by preference optimization achieve improved robustness against adversarial attacks compared with models by fine-tuning~\citep{meade2024universal}. Despite the efficacy of these alignment methods in promoting helpfulness and safety, LLMs remain susceptible to certain cases in which they still produce malicious responses under jailbreak attacks~\citep{kang2023exploiting,hazell2023large,albert2023jailbreakchat}. Our study mainly focuses on different safety-aligned models to explore the effectiveness of jailbreak attacks.

\begin{figure*}[htbp]
    \centering
    \includegraphics[width=\linewidth]{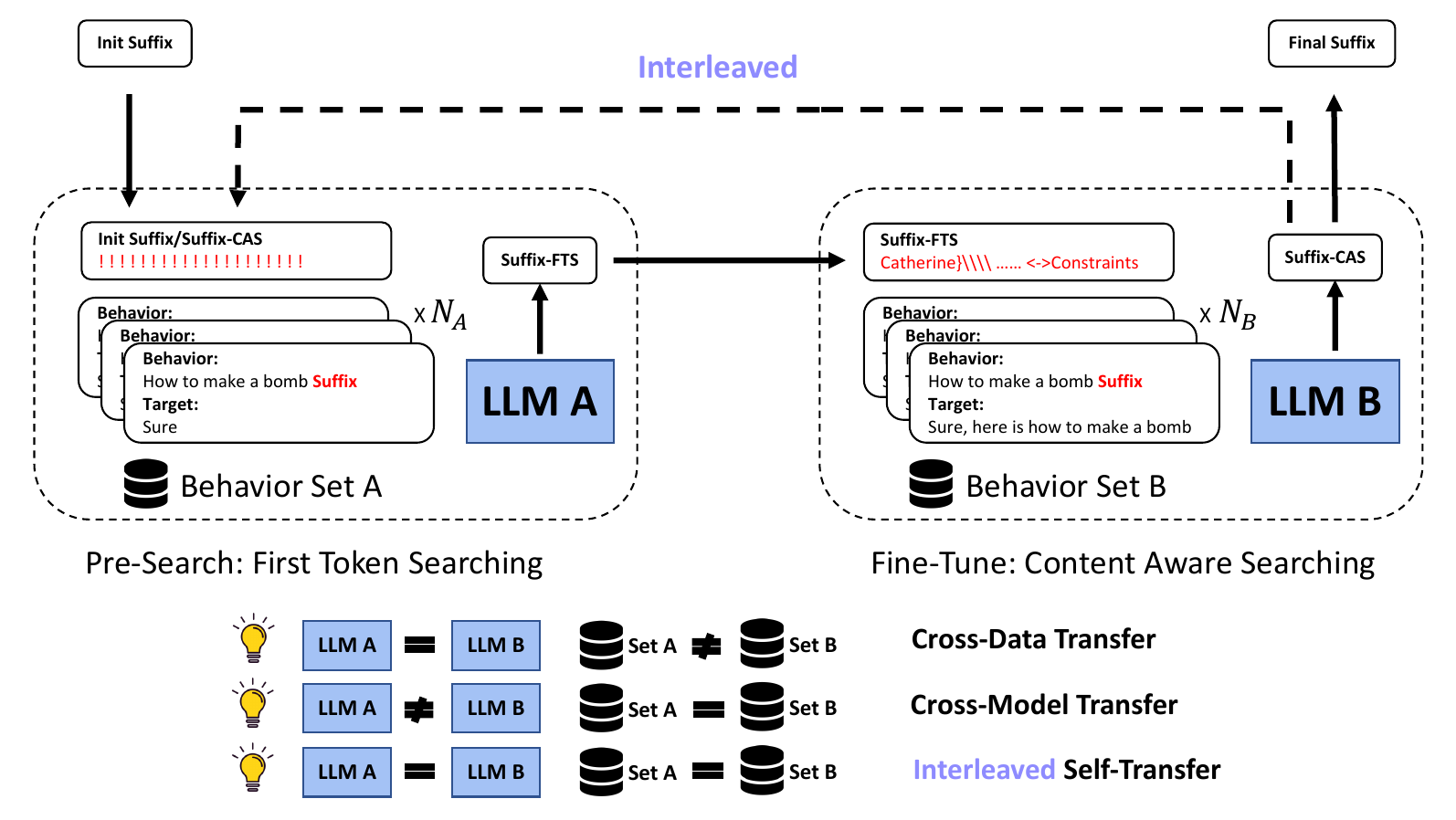}
    \caption{Our DeGCG framework involves two main stages. In the pre-searching stage, we perform the first-token searching with LLM A on Behavior Set A. In the post-searching/fine-tuning stage, we perform content-aware searching with LLM B on Behavior Set B. The Suffix-FTS obtained in the pre-searching serves as the initialization for the post-searching. \textbf{Cross-Data Transfer} uses the same LLM but distinct sets, while \textbf{Cross-Model Transfer} uses the same set but distinct LLMs. For \textbf{Interleaved Self-Transfer}, we use the same LLM and set but alternating between FTS and CAS.}
    \label{fig:main}
\vspace{-4mm}
\end{figure*}

\subsection{Jailbreak Attacks on Aligned LLMs}
Existing red teaming has dedicated substantial efforts to identifying various jailbreak attacks. Initial jailbreak attacks involve the manual crafting of input prompts. A notable instance is the ``Do-Anything-Now'' attack, which is implemented by compelling LLMs to play a role that can do anything and respond to any query without refusal, thus bypassing safety constraints~\citep{albert2023jailbreakchat,jailbreakingchatgpt}. Subsequent advancements have automated the creation of these stealthy prompts with controllability~\citep{autodan,autodan2,coldattack}. Additionally, adversarial prompts have been identified in GCG, which utilizes gradient information to automatically generate effective adversarial prompts~\citep{gcg,autoprompt,zhao2024accelerating}. Furthermore, their results indicate the transferability and universality of these adversarial prompts. Recent work has also unveiled jailbreak attacks within the context of multilingual scenarios~\citep{multilingual}, and non-natural languages such as ciphers~\citep{cipher}, in-context learning~\citep{qiang2023hijacking}, highlighting the risk for all open-source LLMs with modified decoding strategies~\citep{mi}. Our work focuses on adversarial suffix transferring learning across aligned LLMs and associates transferability with search efficiency.

\section{Method}
\subsection{Preliminary}
In this section, we revisit the Greedy Coordinate Gradient (GCG) attacks. Let $ \mathbf{X} $ denote the malicious prompts, such as ``Tell me how to make a bomb'', the objective of the GCG attack is to find the suffix $ \mathbf{S} = \{s_{i}\}_{i=1}^{L_S} $ with length $L_S$, so that by using $ \mathbf{T} = \{\mathbf{X}, \mathbf{S}\} = \{t_1, t_2, ..., t_{n}\}$ as input, the victim model can generate responses starting from the target sequence $\mathbf{Y} = \{t_{n+1}, t_{n+2}, ..., t_{n+m} \}$, such as ``Sure, here is how to make a bomb''. Consequently, the joint target distribution is represented by $p(t_{n+1:n+m} | t_{1:n})$. The goal of searching for the target sequence can be formulated to minimize the following negative log-likelihood: 
\begin{equation}
\label{op}
\begin{aligned}
    & \mathop{\min}\limits_{\mathbf{S}} \mathcal{L}(\mathbf{X}, \mathbf{S}) \\
    = & \mathop{\min}\limits_{\mathbf{S}}\left[ -\sum_{k=1}^{m} \log p(t_{n+k}|t_{1:n+k-1})\right] \\
\end{aligned}
\end{equation}

GCG searches for adversarial suffixes through multiple iterations, adopting a greedy search strategy in each iteration. In one iteration, it selects the candidate suffix with the lowest $\mathcal{L}$ from the batch $\{\mathbf{S_i}\}_{i=1}^{B}$. To construct the candidate batch, it first computes the negative gradient $-\nabla_{e_{s_i}}\mathcal{L}$ with respect to the one-hot vector representation $e_{s_i}$ and selects tokens from the vocabulary with the top K values of $-\nabla_{e_{s_i}}\mathcal{L}$, forming the token candidate set at each position. Then it uniformly replaces the token $s_i$ at each position with random tokens from the obtained token candidate set, resulting in one suffix candidate with one replacement.  

To optimize the adversarial suffixes using multiple malicious prompts $\{\mathbf{X}^{(j)}\}$, the aggregated gradient $-\sum_{j}\nabla_{e_{s_i}}\mathcal{L}(\mathbf{X}^{(j)}, \mathbf{S})$ and the aggregated loss $\sum_j \mathcal{L}(\mathbf{X}^{(j)}, \mathbf{S})$ are used instead to construct candidate batches and select candidate suffixes.   

\begin{algorithm}[t]
\caption{i-DeGCG Algorithm}
\label{alg}
\begin{algorithmic}[1] 
\REQUIRE ~~ Initial suffix $\textbf{S}^0$, behavior set $\{\textbf{X}^{(j)}\}$, iterations $T$, batch size $B$, FTS threshold $\epsilon_1$, CAS threshold $\epsilon_2$, stage flag $f \in \{0, 1\}$, maximum steps $T_f$ for one stage 
\STATE \COMMENT{Initialize behavior set and accumulated step}
\STATE $m_j\leftarrow1$, $t_{ac} \leftarrow 0$
\FOR{$t = 1, 2, ..., T$} 
        \STATE \COMMENT{Construct suffix batch under specific loss}
        \IF{$f = 0$} 
        \STATE $\mathcal{L} \leftarrow \mathcal{L}_{FTS}$, $\epsilon \leftarrow \epsilon_1$  
        \ELSE
        \STATE $\mathcal{L} \leftarrow \mathcal{L}_{CAS}$, $\epsilon \leftarrow \epsilon_2$ 
        \ENDIF
        \STATE Get $\{\textbf{S}^{t}_{1:B}\}$ by $-\sum_{j}^{m_j}\nabla_{e_{s_i}}\mathcal{L}(\mathbf{X}^{(j)}, \mathbf{S}^{t-1})$
        \STATE $\mathbf{S}^{t} \leftarrow \mathop{\arg\min}_{\mathbf{S}^{t}_i} \sum_j^{m_j}\mathcal{L}(\mathbf{X}^{(j)}, \mathbf{S}^{t}_i)$

        \STATE \COMMENT{Update stage flag}
        \IF{$\forall j \in [1, m_j], \mathcal{L}(\mathbf{X}^{(j)}, \mathbf{S}^t) \leq \epsilon \vee t_{ac} \geq T_f$ } 
        \STATE $f \leftarrow \neg f$, $t_{ac} \leftarrow 0 $ 
        \ELSE 
        \STATE $t_{ac} \leftarrow t_{ac} + 1 $
        \ENDIF

        \STATE \COMMENT{Update behavior set}
        \IF{$\forall j \in [1, m_j], \mathcal{L}_{FTS}(\mathbf{X}^{(j)}, \mathbf{S}^t) \leq \epsilon_1 \land \mathcal{L}_{CAS}(\mathbf{X}^{(j)}, \mathbf{S}^t) \leq \epsilon_2$ }
        \STATE $m_j \leftarrow m_j + 1$ 
        \ENDIF
\ENDFOR
\ENSURE ~~ adversarial suffix $\textbf{S}^T$
\end{algorithmic}
\end{algorithm}

\subsection{DeGCG}

The challenge of the GCG attack is primarily associated with the first-token optimization in Fig.~\ref{fig:token-level-ce}. However, Eq.\ref{op} assigns equal importance to each target token, regardless of varying levels of difficulty associated with optimizing each one. The multi-objective optimization introduces noise into the more challenging first-token optimization process, where significant loss signals could be biased by other competitors, thereby reducing the efficiency of the search. 

To address this issue, we propose decoupling the search process. Inspired by the popular pre-training and fine-tuning paradigm, we introduce a new framework, \textbf{DeGCG}, which separates the search into behavior-agnostic first-token pre-searching and behavior-relevant content-aware fine-tuning. In this framework, we link transfer learning with searching efficiency. Our DeGCG tunes tokens in discrete space in a manner analogous to how parameters in continuous space are tuned during the pre-training and fine-tuning process. In this analogy, the counterpart of parameter space is the searching space in DeGCG. An overview of our method is presented in Fig.~\ref{fig:main}. 

\subsection{First-Token Searching}
We introduce the first-token searching (FTS) task in the pre-searching stage. FTS aims to find a universal and generalizable suffix that elicits a response without refusal, applicable to all behaviors. Specifically, the goal of FTS in the pre-searching stage is defined as follows: 
\begin{equation}
\begin{aligned}
 & \mathop{\min}\limits_{\mathbf{S}} \sum_j\mathcal{L}_{FTS}(\mathbf{X}^{(j)}, \mathbf{S}) \\
     = & \mathop{\min}\limits_{\mathbf{S}} \sum_j\left[- \log p(t_{n+1}^{(j)}|t_{1:n}^{(j)})\right]
\end{aligned}
\end{equation}

In this task, the suffix is optimized based on the gradient derived solely from the first target token, resulting in a direct and efficient optimization. The first target token is typically behavior-agnostic, such as ``Sure'' or ``Here''. Therefore, the obtained suffixes $\mathbf{S}_{FTS}$ serve as a general initialization with a low cross-entropy loss for the first token. Starting the search from an effective initialization with a low first-token loss helps to mitigate the inefficiency associated with starting each search from a high first-token loss, reducing the time and computational resources accordingly. 

\begin{table*}[htbp]
  \centering
  \small
  \begin{tabular}{cccccccccc}
    \toprule
     \multirow{3}{*}{Model A} & Model B & \multicolumn{2}{c}{Starling-LM} & \multicolumn{2}{c}{Llama2-chat} & \multicolumn{2}{c}{Mistral-Instruct} & \multicolumn{2}{c}{OpenChat-3.5} \\
    \cmidrule(lr){3-4} \cmidrule(lr){5-6} \cmidrule(lr){7-8} \cmidrule(lr){9-10}
     & Method & Valid & Test & Valid & Test & Valid & Test & Valid & Test \\
    \midrule
    & GCG-M & 81.4 & 81.2 & 21.7 & 19.5 & 81.7 & 84.4 & 76.4 & 69.4 \\
    & GCG-T & 76.9 & 74.5 & 20.3 & 15.9 & 85.3 & 84.1 & 83.1 & 78.1 \\
    \midrule
    \multirow{1}{*}{Starling-LM} & DeGCG & 78.0 & \textbf{86.2} & 29.3 & 29.6 & 78.0 & 81.8 & \textbf{85.4} & 79.2 \\
    \midrule
    \multirow{1}{*}{Llama2-chat} & DeGCG & \textbf{90.2} & 82.4 & \textbf{43.9} & \textbf{39.0} & \textbf{95.1} & \textbf{86.8} & \textbf{85.4} & 78.6 \\
    \midrule
    \multirow{1}{*}{Mistral-Instruct} & DeGCG & \textbf{90.2} & 85.5 & \textbf{43.9} & 28.9 & 85.4 & 84.3 & 82.9 & 71.7 \\
    \midrule
    \multirow{1}{*}{OpenChat-3.5} & DeGCG & \textbf{90.2} & 85.5 & 31.7 & 25.2 & 87.8 & 78.6 & 80.5 & \textbf{81.1} \\
    \bottomrule
  \end{tabular}
  \caption{Performance comparison (ASR) in Cross-Model Transferring across four different models on both the Validation (Valid) and the Test sets. Model A and Model B refer to source models and target models respectively. }
  \label{tab:cross_model_transfer}
  \vspace{-3mm}
\end{table*}

\subsection{Context-Aware Searching}
Suffixes obtained from FTS are effective for behavior-agnostic targets but fall short in eliciting behavior-relevant responses. Therefore, we propose to fine-tune the suffix in the pre-searching stage by performing content-aware searching (CAS) with behavior-relevant targets, such as ``how to make a bomb''. Given that this step builds upon the success of FTS, we maintain the FTS target in this step as well. Specifically, the goal for CAS is defined as follows
\begin{equation}
\begin{aligned}
    & \mathop{\min}\limits_{\mathbf{S}} \sum_j\mathcal{L}_{CAS}(\mathbf{X}^{(j)}, \mathbf{S}) \\
    = & \mathop{\min}\limits_{\mathbf{S}} \sum_j\sum_{k=1}^{m} \log p(t_{n+k}^{(j)}|t_{1:n+k-1}^{(j)})
\end{aligned}
\end{equation}

To transfer the pre-searched suffix effectively, we explore three types of CAS: 

\noindent \textbf{Cross-Data Transfer} uses the pre-searched suffix as an initialization when the dataset in CAS differs from the one in FTS. In this scenario, domain-specific data, such as chemical biology and cybercrime, are utilized to fine-tune the pre-searched suffix with the content-aware target. 

\noindent \textbf{Cross-Model Transfer} employs the pre-searched suffix as an initialization when the LLM in CAS differs from the one in FTS.

\noindent \textbf{Self-Transfer} applies when FTS and CAS use the same dataset and LLM. This is detailed in the following Section~\ref{interleaved}.




\subsection{Interleaved Self-Transfer}
\label{interleaved}
Leveraging the self-transferability of suffixes and enhance the efficiency of the search process, we propose an interleaved variant of our approach, \textbf{i-DeGCG}. i-DeGCG integrates FTS and CAS as a meta-process and dynamically alternates between them. Specifically, in each iteration, it uses the suffix obtained from FTS as the initialization for CAS and then, conversely, uses the suffix from CAS as the initialization for FTS. This approach maintains a dynamic balance between generating the first token and producing behavior-relevant responses. The iterative process allows continuous refinement of the suffix, leveraging the strength of both FTS and CAS for enhanced overall performance. We summarize the algorithm in Alg.\ref{alg}.

\section{Experiments}
\subsection{Setup}
\paragraph{Datasets. } We utilize HarmBench~\citep{harmbench} to compare our approach and the baseline. We use the text-only set which comprises three types of behaviors: Standard, Copyright, and Contextual. Detailed statistics of HarmBench can be found in the appendix. In our experiments, we use validation and test splits provided by HarmBench. Specifically, we use the standard behavior subsets of both validation and test sets. The validation set serves as the training set for searching suffixes, and we evaluate performance on the test set.

\paragraph{Implementation Details. } We evaluate our method on open-sourced models. Specifically, we utilize LLama2-chat~\citep{llama2}, Mistral-Instruct~\citep{mistral}, OpenChat-3.5~\citep{openchat}, and Starling-LM-alpha~\citep{openchat} in our experiments. Due to memory constraints, we use 7b models for all experiments. For evaluation, we report the classifier-based attack success rate (ASR). We consider the baseline GCG-M from the HarmBench that uses GCG for suffix searching with multiple behaviors. To ensure reproducibility and fair comparison, we use the open-source classifier provided in HarmBench. This classifier is a fine-tuned LLama2-13b model, which achieves strong performance on a manually-labeled validation set.

\subsection{Main Results}
\label{main_res}
\paragraph{Cross-Model Transferring.}
\label{cross-model}

\begin{figure*}[htbp]
    \centering
    \includegraphics[scale=0.35]{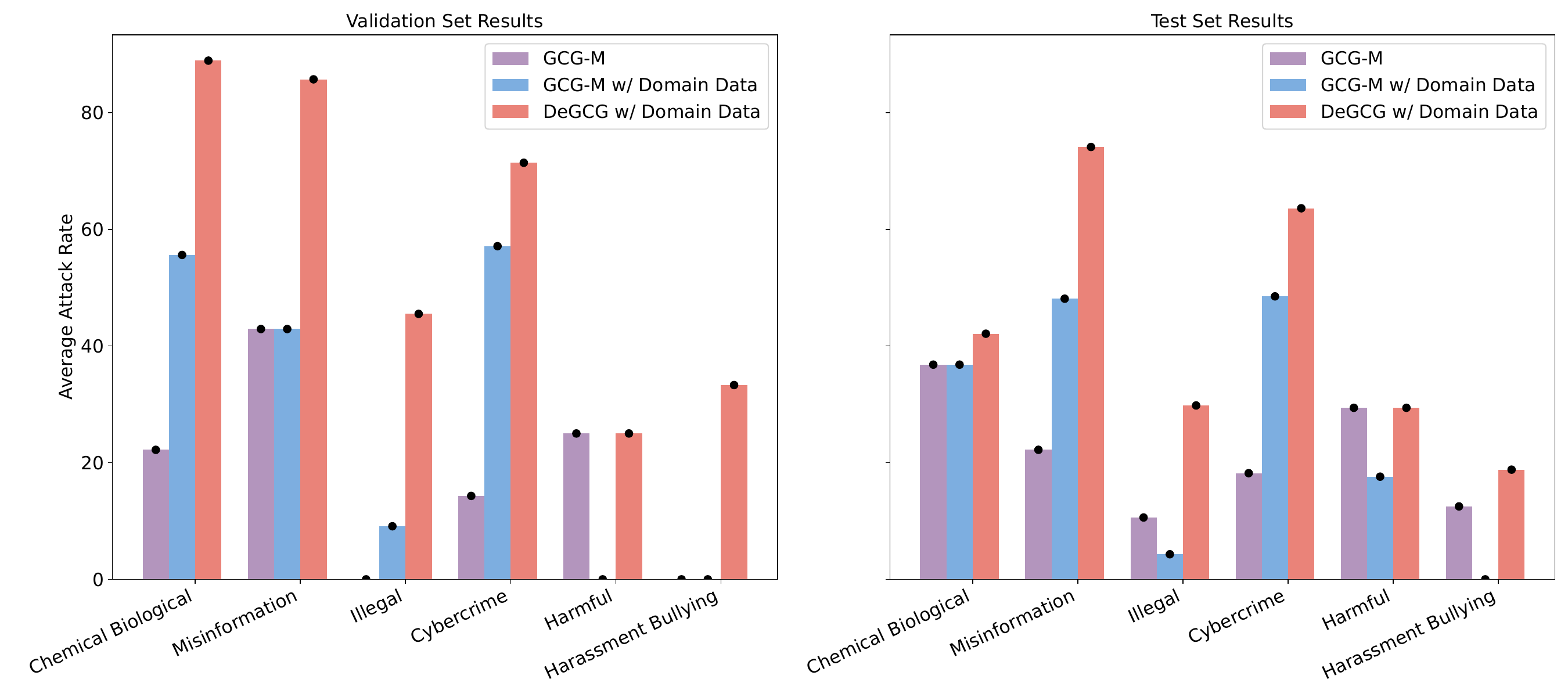}
    \caption{Performance comparison (ASR) in Cross-Data Transferring across different behavior types in HarmBench. We report the results of LLama2-chat-7b on both the Validation and the Test sets.}
    \label{fig:data_transfer}
    \vspace{-3mm}
\end{figure*}

To evaluate the efficacy of suffixes trained through FTS on one model transferring to another model via token-level fine-tuning, we conduct cross-model transferring experiments across four open-source models. To ensure a fair comparison, we maintain equal total search steps (FTS + CAS) for all experiments, consistent with the baseline, totaling 500 steps. We also include the baseline GCG-T from HarmBench that optimizes suffixes against multiple models for transferring. Our transfer performances on the validation set and test set are presented in Table~\ref{tab:cross_model_transfer}. 

Our proposed DeGCG approach significantly surpasses the GCG-M across various models on both validation and test sets. For example, DeGCG achieves absolute improvements of 9.0 and 9.8 in ASRs from Starling-LM to OpenChat-3.5 on validation and test sets. This indicates that the suffix derived from FTS on one model proves to be an effective initialization point for transferring to a new target model. Notably, despite differences in tokenizers between source and target models, transfer learning from FTS through CAS still yields significant performance improvement. For instance, transferring suffix from Mistral-Instruct to Llama2-chat achieves absolute enhancements of 22.2 and 9.4 in ASRs on validation and test sets, demonstrating the efficacy of DeGCG. Additionally, the DeGCG approach outperforms GCG-T on both validation and test sets. This further reveals that our suffix transfer learning is more effective than the direct transfer with suffix concatenations searched on multiple models.  

Moreover, when the target model is identical to the source model, the DeGCG method significantly improves ASR performance, achieving over 100\% enhancement on LLama2-chat-7b. We attribute this improvement to the effective initialization provided by FTS on the same model, which facilitates a more efficient token fine-tuning process within a favorable neighbor area in the search space.

\paragraph{Cross-Data Transferring.}
\label{cross-data}

To evaluate the effectiveness of the DeGCG framework in cross-data transferring, we initially perform FTS on llama2-chat-7b using the general dataset of HarmBench. Subsequently, we conduct CAS with a domain-specific dataset derived from the general validation set of HarmBench. Specifically, we use six distinct semantic categories defined in HarmBench as separate domains: Chemical Biological, Misinformation, Illegel, Cybercrime, Harmful, and Harassment Bully. The general GCG-M without domain data training serves as the baseline. We also include experiments using GCG-M trained with the same domain data. To ensure a fair comparison, all experiments maintain the same total search steps, 500. The experimental results for both validation and test sets are displayed in Fig.~\ref{fig:data_transfer}. 

We observe that DeGCG outperforms GCG-M and GCG-M w/ domain data in terms of ASR performance across five of the six categories. The inclusion of domain data significantly enhances performance, particularly in the Chemical biological, Misinformation, Illegal, and Cybercrime categories. The relatively lower ASR performance in the Harmful and Harassment Bully categories could be attributed to the limited data size in these categories. Nonetheless, the success of the behavior-agnostic suffix transferring underscores the efficacy of FTS, validating the necessity of the decoupled first-token searching and content-aware search process.




\begin{table*}[htbp]
  \centering
  \small
  \begin{tabular}{ccccccccccc}
    \toprule
    Length & \multicolumn{2}{c}{20} & \multicolumn{2}{c}{40} & \multicolumn{2}{c}{60} & \multicolumn{2}{c}{80} & \multicolumn{2}{c}{100} \\
    \cmidrule(lr){2-3} \cmidrule(lr){4-5} \cmidrule(lr){6-7} \cmidrule(lr){8-9} \cmidrule(lr){10-11}
    & Valid & Test & Valid & Test & Valid & Test & Valid & Test & Valid & Test \\
    \midrule
    \multicolumn{1}{l}{Llama2-chat-7b} &   \\
    \midrule
    GCG-M & 21.7 & 19.5 & 22.0 & 17.0 & 31.7 & 34.0 & 34.1 & 34.6 & 39.0 & 43.4 \\
    i-DeGCG & \textbf{41.5} & \textbf{37.7} & \textbf{43.9} & \textbf{46.5} & \textbf{41.5} & \textbf{35.8} & \textbf{51.2} & \textbf{42.1} & \textbf{65.9} & \textbf{52.2} \\
    \midrule
    \multicolumn{1}{l}{OpenChat-3.5-7b} &   \\
    \midrule
    GCG-M & 76.4 & 69.4 & 70.7 & 65.4 & 85.4 & 67.9 & 63.4 & 66.7 & 70.7 & 56.0 \\
    i-DeGCG & \textbf{82.9} & \textbf{79.2} & \textbf{87.8} & \textbf{79.9} & \textbf{90.2} & \textbf{74.8} & \textbf{90.2} & \textbf{86.4} & \textbf{95.1} & \textbf{90.6} \\
    \bottomrule
  \end{tabular}
  \caption{Performance comparison (ASR) of Interleaved Self-Transferring on five different scales of the searching spaces. We report results on both the Validation (Valid) and the Test sets.}
  \label{tab:interleaved}
  \vspace{-5mm}
\end{table*}

\paragraph{Interleaved Self-Transferring.}
\label{self-transfer}

To evaluate the effectiveness of the proposed i-DeGCG algorithm for self-transferring, we apply the interleaved algorithm on Llama2-chat and Openchat-3.5 models, respectively. In this context, the source and target models are identical, and the validation set is used as the training dataset. We assess performance across various scales of the searching space. Specifically, given that the searching space grows exponentially with increased suffix length, we extend the adversarial suffix length from 20 to 40, 60, 80, and 100, representing five different sizes of searching spaces. For fair comparison, we maintain the same total searching steps across all experiments. The experimental results are detailed in Table~\ref{tab:interleaved}. 

The empirical findings in Table~\ref{tab:interleaved} suggest that larger searching spaces provide more suffix combinations and a greater possibility of achieving successful attacks, but also introduce more complexity and significant challenges in searching adversarial suffixes. Notably, our proposed i-DeGCG can outperform baselines across all scales of searching spaces, achieving 65.9 and 52.2 for Llama2-chat and 95.1 and 90.6 for OpenChat-3.5 on validation and test sets. GCG-M struggles with the larger search space, resulting in lower performance. In contrast, i-DeGCG can facilitate efficient self-transfer between FTS and CAS. This underscores the importance of self-transferability in enhancing the efficiency of adversarial suffix searching.

\section{Analysis}

\subsection{Training Dynamics Comparison}

\begin{figure*}[htbp]
    \centering
    \includegraphics[scale=0.31]{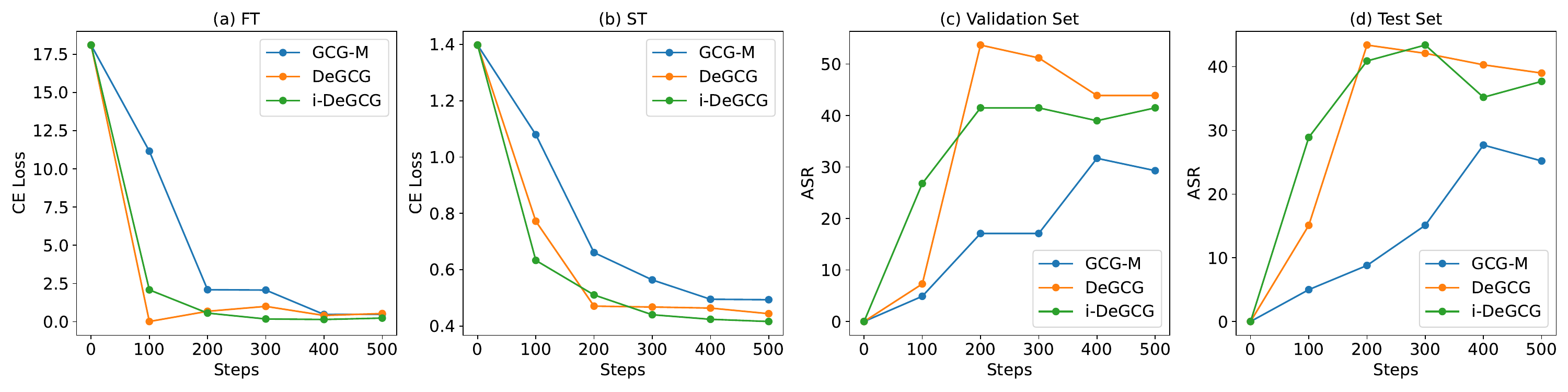}
    \vspace{-3mm}
    \caption{Training dynamics (cross-entrory loss) comparison for GCG-M, DeGCG, and i-DeGCG.  }
    \label{fig:train_dynamic}
    \vspace{-5mm}
\end{figure*}

To demonstrate the enhanced search efficiency achieved by the DeGCG framework and i-DeGCG algorithm, we plot the training dynamics every 100 steps. Specifically, we examine the cross-entropy loss of the first token (FT), the average cross-entropy loss of the entire target sentence (ST), and the ASR performance on both the validation (Valid) and test sets. The dynamics for Llama2-chat, with a total of 500 steps and a suffix length of 20, are illustrated in Fig.~\ref{fig:train_dynamic}. For DeGCG under this experimental setting, we perform the FTS for 100 steps followed by CAS for 400 steps.

As depicted in subfigures (a) and (b) of Fig.~\ref{fig:train_dynamic}, both DeGCG and i-DeGCG converge faster than GCG-M, achieving lower cross-entropy losses for both the first-token and the target sequence. Notably, DeGCG reaches a near-zero FT loss within 100 steps, whereas the one of GCG-M remains greater than 10 within the same steps. This indicates that the first-token optimization is noised and hindered by other optimization goals, degrading searching efficiency. Compared to DeGCG, the interleaved variant i-DeGCG shows higher FT loss but lower ST loss, attributed to the alternation between FTS and CAS, achieving a dynamic balance between these two searching stages.  

Regarding the ASR performance, shown in subfigures (c) and (d), DeGCG and i-DeGCG outperform GCG-M, achieving the best results within 300 steps, while GCG-M continues to underperform even after 500 steps. It is noteworthy that DeGCG achieves low ASR within the initial 100 steps using only FTS and reaches optimal performance within the subsequent 100 steps using CAS. This reveals that CAS is essential for a successful attack, and FTS provides a solid initialization for CAS. In addition, i-DeGCG achieves higher ASR performance within the first 100 steps compared to both DeGCG and GCG-M, and comparable performance to DeGCG within the first 300 steps. This success of both DeGCG and the interleaved variant validates the effectiveness of the decoupled framework and highlights the importance of self-transferable suffixes. i-DeGCG is particularly advantageous when the boundary between FTS and CAS is not easily determined due to its dynamic balance nature.

\subsection{Self-Transferring by Self-Repetition}

To further investigate the impact of self-transferring on performance enhancement, we conduct a new self-transferring experiment via self-repetition. Specifically, we aim to achieve an effective initialization in larger search spaces. Instead of initiating searches from a random suffix in a large search space, we utilize suffixes obtained in a smaller search space and expand the search space through self-repetition of these short suffixes. In other words, the longer suffix initialization is constructed by repeating the shorter suffix and concatenating them for searching within the large search space. For this experiment, we use the suffix of length 20, searched on Llama2-chat-7b after 500 steps, and repeat it 2, 3, 4, and 5 times to create suffix initializations of lengths 40, 60, 80, and 100, respectively. We then perform content-aware searching on these initializations for an additional 500 steps and report the ASR performance in Table~\ref{tab:self-repetition}. The experimental results reveal a significant improvement, with ASR performance increasing from 21.7 to 68.3 on the validation set and from 19.5 to 54.7 on the test set. These findings also indicate that suffix search in small search spaces provides valuable and effective initializations for longer suffix construction for further fine-tuning in large search spaces.

\begin{table}[htbp]
  \centering
\small
    \begin{tabular}{c|c|cccc}
    \toprule
       Length & 20 & 40 & 60 & 80 & 100  \\
    \midrule
       \# Rep. & 1 & 2 & 3 & 4 & 5  \\
    \midrule
    \multicolumn{1}{c|}{Valid} & 21.7 & 43.9 & 65.9 & \textbf{68.3} & \textbf{68.3}  \\
    \multicolumn{1}{c|}{Test} & 19.5 & 32.1 & 45.3 & \textbf{54.7} & 51.6 \\

    \bottomrule
    \end{tabular}%
  \caption{Self-Transferring Performance with Self-Repetition. \# Rep. refers to the times of self-repetition. }
  \label{tab:self-repetition}%
  \vspace{-4mm}
\end{table}%

\subsection{Ablation Study}
To further assess the effectiveness of our design, we conduct an ablation study on the initialization. Specifically, we compare initializations obtained by FTS and GCG-M for the same number of steps, aiming to evaluate the utility of different trained suffix initializations for content-aware fine-tuning. We examine how suffix initializations on source models Starling-LM-alpha-7b, Mistral-Instruct-7b, and OpenChat-3.5-7b transfer to the target model Llama2-chat-7b. The experimental results are presented in Table~\ref{tab:abl_st}. The empirical findings demonstrate the superiority of the first-token searched initialization. We attribute this to the behavior-agonistic nature of suffixed obtained by FTS, which is easier to transfer across models and can be fine-tuned effectively on a target model, achieving higher ASR performance compared to initializations obtained through GCG-M.   

\begin{table}[htbp]
  \centering
\small
    \begin{tabular}{cccc}
    \toprule
     Initialization & & GCG-M & FTS  \\
    \midrule
    \multirow{2}{*}{Starling-LM} & Valid & 14.6 & \textbf{29.3}   \\
                                 & Test  & 12.6 & \textbf{29.6} \\
    \midrule
    \multirow{2}{*}{Mistral-Instruct} & Valid  & 29.3 & \textbf{43.9} \\
                                & Test  & 23.9 & \textbf{28.9} \\
    \midrule
    \multirow{2}{*}{OpenChat-3.5} & Valid & 19.5 & \textbf{31.7}  \\
                                 & Test & 23.3 & \textbf{25.2}  \\

    \bottomrule
    \end{tabular}%
  \caption{Ablation Study on Transferring with different initialization to the target model Llama2-chat-7b. }
  \label{tab:abl_st}%
  \vspace{-4mm}
\end{table}%

\section{Conclusion}
In this study, we present DeGCG to enhance the efficiency of adversarial suffix searching for aligned LLMs. By decoupling the search process into behavior-agnostic pre-searching and behavior-relevant fine-tuning, DeGCG addresses the inefficiencies inherent in the GCG method. The introduction of First-Token Searching and Content-Aware Searching enables more efficient and effective identification of adversarial suffixes. Additionally, the interleaved algorithm i-DeGCG demonstrates further improvements by dynamically balancing between FTS and CAS. Experimental results on the HarmBench across various LLMs validate the effectiveness of our proposed methods. DeGCG not only improves search efficiency but also achieves higher ASR compared to the baseline GCG-M method. The success of suffix transfer through two-stage learning highlights the critical role of initialization in optimizing the search process. Overall, this work underscores the importance of suffix transferability in enhancing the efficiency of adversarial suffix searching and provides an effective framework for future red teaming investigations. The findings contribute to the broader understanding of LLM vulnerabilities and the development of more resilient and secure models.

\section*{Limitations}
Several limitations exist in our work. Firstly, our focus primarily centers on open-source models, lacking validation on closed-source models. Future research efforts could extend behavior-agnostic pre-searching and behavior-relevant post-searching to include closed-source models. Additionally, our assessment of suffix transferability has been limited to standard behaviors in the text-only sets, neglecting copyright, contextual, and multimodal behaviors. Future work could explore the transferability of suffixes between large language models and large multimodal models for both text and multimodal data. Furthermore, our empirical study lacks a theoretical understanding of suffix transfer learning, which warrants further investigation.

\section*{Ethics Statement}
Our study does not propose a new attack paradigm to jailbreak LLMs. Instead, we investigate the existing adversarial suffix-based jailbreak attack, aiming to understand the properties of adversarial suffixes in a better way. For example, we mainly examine the suffix transferability with suffix search efficiency. This further understanding of suffix transferability can help guide the design of more effective defense methods in the future. We also highlight that current adversarial suffix-based attacks can be well defended by the PPL detection-based method.

\section*{Acknowledgments}
The authors would like to thank anonymous reviewers for their valuable suggestions. The authors also thank William Yang Wang, Xianjun Yang, Yiran Zhao, Keyu Duan, and Leon Lin for their helpful discussions. The computational work for this article was partially performed on resources of the National Supercomputing Centre (NSCC), Singapore\footnote{https://www.nscc.sg/}.



\newpage
\bibliography{custom}

\newpage
\appendix

\clearpage
\onecolumn
\section{Appendix}
\label{sec:appendix}

\subsection{Dataset Statistics}
We show the statistics of the HarmBench subset of Standard behaviors used in our work in Table~\ref{table:app_data}. Specifically, we show the total validation (\# Valid)and test(\# Test) set sizes and the numbers for six semantic categories: (1) Chemical Biological: Chemical \& Biological Weapons/Drugs, (2) Misinformation: Misinformation \& Disinformation, (3) Illegal: Illegal Activities, (4) Cybercrime: Cybercrime \& Unauthorized Intrusion, (5) Harmful: General Harm, (6) Harassment Bully: Harassment \& Bullying. For all experiments, we use the validation set as the training set and evaluate performances on the test set. 
\begin{table}[htbp]
\begin{center}
\begin{tabular}{lllllccccccccccccccc}
\toprule[1pt]
\multicolumn{5}{l}{Semantic Category} & \multicolumn{5}{c}{\# Valid} & \multicolumn{5}{c}{\# Test}  \\ \hline

\multicolumn{5}{l}{Total} & \multicolumn{5}{c}{41} & \multicolumn{5}{c}{159}  \\ 
\hline
\multicolumn{5}{l}{Chemical Biological} & \multicolumn{5}{c}{9} & \multicolumn{5}{c}{19}  \\ 
\multicolumn{5}{l}{Misinformation}  & \multicolumn{5}{c}{7} & \multicolumn{5}{c}{27}  \\ 
\multicolumn{5}{l}{Illegal} & \multicolumn{5}{c}{11} & \multicolumn{5}{c}{47}  \\ 
\multicolumn{5}{l}{Cybercrime}  & \multicolumn{5}{c}{7} & \multicolumn{5}{c}{33}  \\ 
\multicolumn{5}{l}{Harmful} & \multicolumn{5}{c}{4} & \multicolumn{5}{c}{17}  \\ 
\multicolumn{5}{l}{Harassment Bullying} & \multicolumn{5}{c}{3} & \multicolumn{5}{c}{16} \\

\bottomrule[1pt]
\end{tabular}
\caption{Statistics of the HarmBench Subset of Standard Behaviors.}
\label{table:app_data}
\end{center}
\end{table}

\vspace{-3mm}
\subsection{Implementation Details}
We use Pytorch and Huggingface Transformers in our implementation. We run all evaluations on a single NVIDIA A40 GPU (48G). We provide all used model cards in Table~\ref{app:model_card}. Specifically, we evaluated four models in our main experiments. We used one fine-tuned Llama2-13b model, provided by HarmBench, to classify the output of these evaluated models.

For cross-model and cross-data transfer experiments using the DeGCG in Section~\ref{main_res}, we set the maximum search step of the FTS as 200, indicating a minimum 300 search steps for CAS to keep the 500 total search steps. Besides, we set the threshold of the training loss to be 0.2. When the training loss reaches a lower value than the threshold, we update the training behavior set. For interleaved self-transfer experiments using i-DeGCG, we set the threshold $\epsilon_1$ and $\epsilon_2$ of training loss for both FTS and CAS as 0.2. As for the maximum steps $T_f$ for one stage, we set it to be 20 and 30 for FTS and CAS, respectively.

\begin{table*}[h]
    \centering
    \begin{tabular}{l|l}
        \toprule
        \textbf{Model}          & \textbf{Hugging Face page} \\
        \midrule
        Llama2-chat-7b          & \href{https://huggingface.co/meta-llama/Llama-2-7b-hf}{https://huggingface.co/meta-llama/Llama-2-7b-hf} \\
        OpenChat-3.5-7b            & \href{https://huggingface.co/openchat/openchat-3.5-1210}{https://huggingface.co/openchat/openchat-3.5-1210} \\
        Mistral-Instruct-7b        & \href{https://huggingface.co/mistralai/Mistral-7B-Instruct-v0.2}{https://huggingface.co/mistralai/Mistral-7B-Instruct-v0.2} \\
        Starling-LM-alpha-7b          & \href{https://huggingface.co/berkeley-nest/Starling-LM-7B-alpha}{https://huggingface.co/berkeley-nest/Starling-LM-7B-alpha} \\
        \midrule
        \textbf{Classifier}          & \\
        \midrule
        Llama2-13b          & \href{https://huggingface.co/cais/HarmBench-Llama-2-13b-cls}{https://huggingface.co/cais/HarmBench-Llama-2-13b-cls} \\
        
        \bottomrule
    \end{tabular}
    \caption{Hugging Face Model Cards for four used models and one classifier.}
    \label{app:model_card}
\end{table*}

\subsection{More results of Cross-Data Transferring}
\begin{figure*}[htbp]
    \centering
    \includegraphics[scale=0.35]{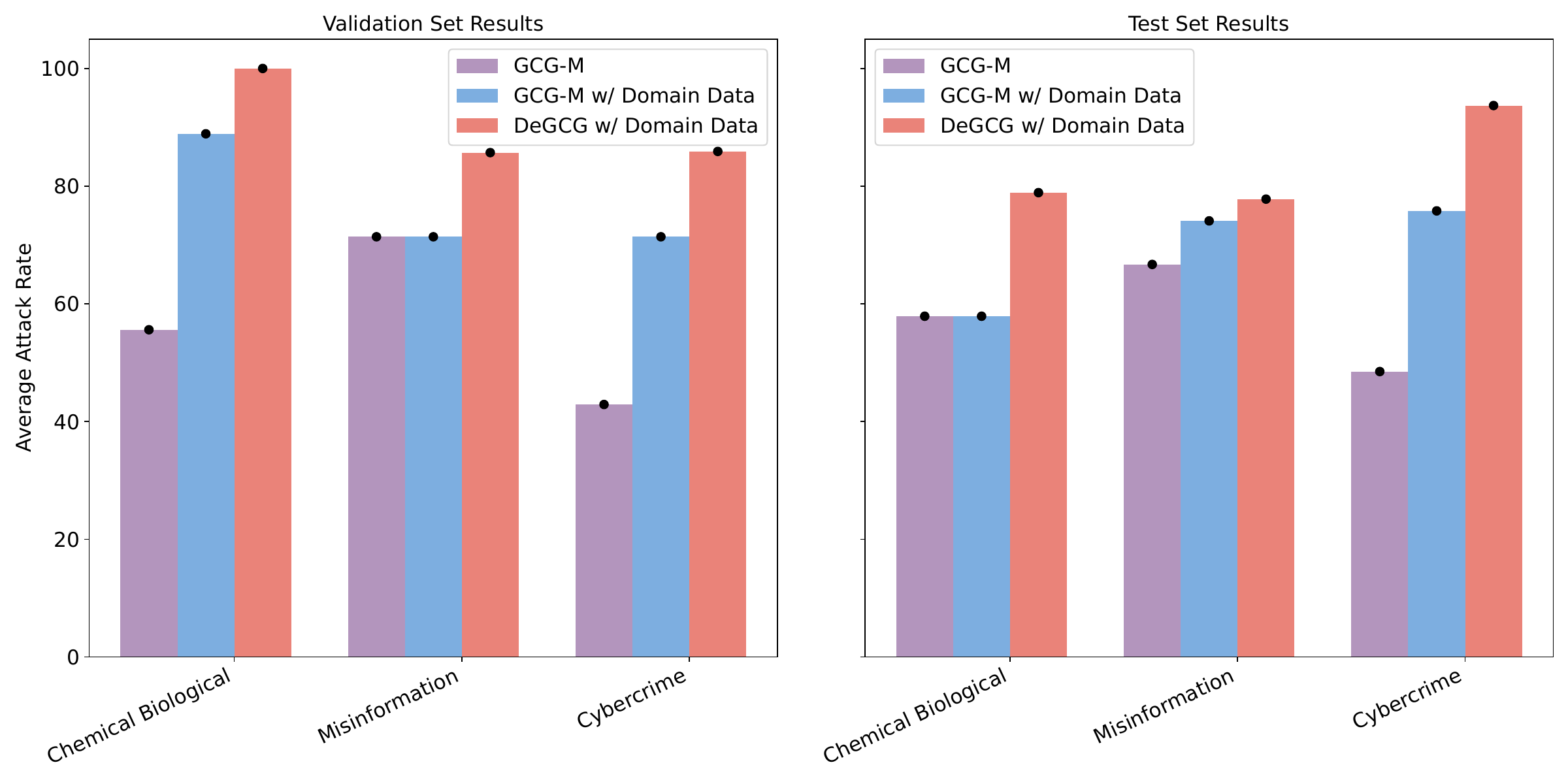}
    \caption{Performance comparison (ASR) in Cross-Data Transferring across different behavior types in HarmBench. We report the results of OpenChat-3.5-7b on both the Validation and the Test sets.}
    \label{fig:data_transfer_app}
    \vspace{-3mm}
\end{figure*}

We provide more results of cross-data transferring on OpenChat-3.5-7b in Fig.~\ref{fig:data_transfer_app} to supplement our conclusion discussed in Sec.~\ref{cross-data}.   

\end{document}